\documentclass[11pt,a4paper]{article}
\usepackage[hyperref]{acl2019}

\usepackage{amsmath,amsfonts,bm}

\def\eqref#1{equation~\ref{#1}}

\def\1{\bm{1}}

\def\vb{{\bm{b}}}

\def\vv{{\bm{v}}}

\def\vx{{\bm{x}}}
\def\vy{{\bm{y}}}
\def\vz{{\bm{z}}}

\def\mI{{\bm{I}}}

\def\mU{{\bm{U}}}

\def\mW{{\bm{W}}}

\DeclareMathAlphabet{\mathsfit}{\encodingdefault}{\sfdefault}{m}{sl}
\SetMathAlphabet{\mathsfit}{bold}{\encodingdefault}{\sfdefault}{bx}{n}

\def\sR{{\mathbb{R}}}

\usepackage{times}
\usepackage{latexsym}

\usepackage{hyperref}
\usepackage{url}

\usepackage{caption}

\usepackage{tabularx}
\usepackage{amsmath,amssymb,amsthm}
\usepackage{graphicx}
\usepackage{subcaption}
\usepackage{listings}
\usepackage{upquote}
\usepackage{wrapfig}
\usepackage{lipsum}
\usepackage{array}
\usepackage{multicol}
\usepackage{multirow}
\usepackage{url}
\usepackage{color}
\usepackage{wrapfig,lipsum,booktabs}
\usepackage{makecell}
\usepackage{booktabs}
\usepackage{changepage}
\usepackage{xcolor}

\usepackage{algorithm}
\usepackage[noend]{algpseudocode}
\usepackage{tabulary}
\newcolumntype{K}[1]{>{\centering\arraybackslash}p{#1}}

\usepackage{tabu}
\usepackage{arydshln}

\aclfinalcopy 

\title{Exploiting Invertible Decoders \\for Unsupervised Sentence Representation Learning}

\author{Shuai Tang \hspace{1cm} Virginia R. de Sa \\
  Department of Cognitive Science,  Hal\i c\i o\u{g}lu Data Science Institute, UC San Diego \\
  {\tt \{\href{mailto:shuatang93@ucsd.edu}{shuaitang93},\href{{mailto:desa@ucsd.edu}}{desa}\}@ucsd.edu} 
  }

\date{}

\begin{document}
\maketitle
\begin{abstract}
Encoder-decoder models for unsupervised sentence representation learning using the distributional hypothesis effectively constrain the learnt representation of a sentence to only that needed to reproduce the next sentence. While the decoder is important to constrain the representation, these models tend to discard the decoder after training since only the encoder is needed to map the input sentence into a vector representation. However, parameters learnt in the decoder also contain useful information about the language. In order to utilise the decoder after learning, we present two types of decoding functions whose inverse can be easily derived without expensive inverse calculation. Therefore, the inverse of the decoding function serves as another encoder that produces sentence representations. We show that, with careful design of the decoding functions, the model learns good sentence representations, and the ensemble of the representations produced from the encoder and the inverse of the decoder demonstrate even better generalisation ability and solid transferability.
\end{abstract}

\section{Introduction}
Learning sentence representations from unlabelled data is becoming increasingly 
prevalent in both the machine learning and natural language processing research communities, as it efficiently and cheaply allows knowledge extraction that can successfully transfer to downstream tasks.
Methods built upon the distributional hypothesis \cite{harris1954distributional} 
and distributional similarity \cite{firth57synopsis} can be roughly categorised into two types:

\textbf{Word-prediction Objective:} The objective pushes the system to make better predictions of words in a given sentence. As the nature of the objective is to predict words, these are also called generative models. In one of the two classes of models of this type, an encoder-decoder model is learnt using a corpus of contiguous sentences \cite{Kiros2015SkipThoughtV,Gan2017LearningGS,Tang2018SpeedingUC} to make predictions of the words in the next sentence given the words in the current one. After training, the decoder is usually discarded as it is only needed during training and is not designed to produce sentence representations. In the other class of models of this type, a large language model is learnt \cite{Peters2018DeepCW,Radford2018Improving,Devlin2018BERTPO} on unlabelled corpora, which could be an autoregressive model or a masked language model, which gives extremely powerful language encoders but requires massive computing resources and training time.



\textbf{Similarity-based Objective:} The objective here relies on a predefined similarity function to enforce the model to produce more similar representations for adjacent sentences than those that are not \cite{Li2014AMO,Jernite2017DiscourseBasedOF,Nie2017DisSentSR,logeswaran2018an}. Therefore, the inductive biases introduced by the two key components, the differential similarity function and the context window, in the objective crucially determine the quality of learnt representations and what information of sentences can be encoded in them.


To avoid tuning the inductive biases in the similarity-based objective, we follow the word-prediction objective with an encoder and a decoder, and we are particularly interested in exploiting invertible decoding functions, which can then be used as additional encoders during testing. The contribution of our work is summarised as follows:


\begin{enumerate}
    \item The decoder is used in testing to produce sentence representations. With careful design, the inverse function of the decoder is easy to derive with no expensive inverse calculation.
    \item The inverse of the decoder provides high-quality sentence representations as well as the encoder does, and since the inverse function of the decoder naturally behaves differently from the encoder; thus the representations from both functions complement each other and an ensemble of both provides good results on downstream tasks.
    \item The analyses show that the effectiveness of the invertible constraint enforced on the decoder side and learning from unlabelled corpora helps the produced representations to better capture the meaning of sentences.

\end{enumerate}

\section{Related Work}
Learning vector representations for words with a word embedding matrix as the encoder and a context word embedding matrix as the decoder \cite{Mikolov2013EfficientEO,Lebret2014WordET,Pennington2014GloveGV,Bojanowski2017EnrichingWV} can be considered as a word-level example of our approach, as the models learn to predict the surrounding words in the context given the current word, and the context word embeddings can also be utilised to augment the word embeddings \cite{Pennington2014GloveGV,Levy2015ImprovingDS}. 
We are thus motivated to explore the use of sentence decoders after learning instead of ignoring them as most sentence encoder-decoder models do. 

Our approach is to invert the decoding function in order to use it as another encoder to assist the original encoder. In order to make computation of the inverse function well-posed and tractable, careful design of the decoder is needed. A simple instance of an invertible decoder is a linear projection with an orthonormal square matrix, whose transpose is its inverse. A family of bijective transformations with non-linear functions \cite{Dinh2014NICENI,Rezende2015VariationalIW,Kingma2016ImprovingVI} can also be considered as it empowers the decoder to learn a complex data distribution.

In our paper, we exploit two types of plausible decoding functions, including linear projection and bijective functions with neural networks \cite{Dinh2014NICENI}, and with proper design, the inverse of each of the decoding functions can be derived without expensive inverse calculation after learning. Thus, the decoder function can be utilised along with the encoder for building sentence representations. We show that the ensemble of the encoder and the inverse of the decoder outperforms each of them.

\section{Model Design}
\label{sec:model}
Our model has similar structure to that of skip-thought \cite{Kiros2015SkipThoughtV} and, given the neighbourhood hypothesis \cite{Tang2017Rethinking}, learns to decode the next sentence given the current one instead of predicting both the previous sentence and the next one at the same time.

\subsection{Training Objective}
Given the finding \cite{Tang2018SpeedingUC} that neither an autoregressive nor an RNN decoder is necessary for learning sentence representations that excel on downstream tasks as the autoregressive decoders are slow to train  and the quality of the generated sequences is not highly correlated with that of the representations of the sentences, our model only learns to predict words in the next sentence in a non-autoregressive fashion. 

Suppose that the $i$-th sentence $S_i=\{w_1,w_2,...,w_{N_i}\}$ has $N_i$ words, and $S_{i+1}$ has $N_{i+1}$ words. The learning objective is to maximise the averaged log-likelihood for all sentence pairs:
\begin{align}
\ell_{S_{i+i}|S_i}(\boldsymbol{\phi},\boldsymbol{\theta})=\frac{1}{N_{i+1}}\sum_{w_j\in S_{i+1}}\log P(w_j|S_i) \label{og} \nonumber
\end{align}
where $\boldsymbol{\theta}$ and $\boldsymbol{\phi}$ contain the parameters in the encoder $f_\text{en}(S_i;\boldsymbol{\theta})$ and the decoder $f_\text{de}(\vz_i;\boldsymbol{\phi})$ respectively. The forward computation of our model for a given sentence pair $\{S_i, S_{i+1}\}$, in which the words in $S_i$ are the input to the learning system and the words in $S_{i+1}$ are targets is defined as:
\begin{align}
    \vz_i &= f_\text{en}(S_i;\boldsymbol{\theta}) \nonumber \\
    \vx_i &= f_\text{de}(\vz_i;\boldsymbol{\phi}) \nonumber
\end{align}
where $\vz_i$ is the vector representation of $S_i$, and $\vx_i$ is the vector output of the decoder which will be compared with the vector representations of words in the next sentence $S_{i+1}$. Since calculating the likelihood of generating each word involves a computationally demanding softmax function, the negative sampling method \cite{Mikolov2013EfficientEO} is applied to replace the softmax, and $\log P(w_j|s_i)$ is calculated as:
\begin{align}
\log\sigma(\vx_i^\top \vv_{w_j}) + \sum_{k=1}^{K}\mathbb{E}_{w_k\sim P_e(w)}\log\sigma(-\vx_i^\top \vv_{w_k})  \nonumber
\end{align}
where $\vv_{w_k}\in\sR^{d_\vv}$ is the pretrained vector representation for $w_k$, the empirical distribution $P_e(w)$ is the unigram distribution of words in the training corpus raised to power 0.75 as suggested in the prior work \cite{Mikolov2013DistributedRO}, and $K$ is the number of negative samples. In this case, we enforce the output of the decoder $\vx_i$ to have the same dimensionality as the pretrained word vectors $\vv_{w_j}$. The loss function is summed over all contiguous sentence pairs in the training corpus. For simplicity, we omit the subscription for indexing the sentences in the following sections.

\subsection{Encoder}
The encoder $f_\text{en}(S;\boldsymbol{\theta})$ is a bi-directional Gated Recurrent Unit \cite{Chung2014EmpiricalEO} with $d$-dimensions in each direction. It processes word vectors in an input sentence $\{\vv_{w_1},\vv_{w_2},...,\vv_{w_{N}}\}$ sequentially according to the temporal order of the words, and generates a sequence of hidden states. During learning, in order to reduce the computation load, only the last hidden state serves as the sentence representation $\vz\in\sR^{d_\vz}$, where $d_\vz=2d$.

\subsection{Decoder}
As the goal is to reuse the decoding function $f_{\text{de}}(\vz)$ as another plausible encoder for building sentence representations after learning rather than ignoring it, one possible solution is to find the inverse function of the decoder function during testing, which is noted as $f^{-1}_{\text{de}}(\vx)$. In order to reduce the complexity and the running time during both training and testing, the decoding function $f_{\text{de}}(\vz)$ needs to be easily invertible. Here, two types of decoding functions are considered and explored.

\subsubsection{Linear Projection}
In this case, the decoding function is a linear projection, which is $\vx = f_{\text{de}}(\vz)=\mW\vz + \vb$, where $\mW\in\sR^{d_\vv\times d_\vz}$ is a trainable weight matrix and $\vb\in\sR^{d_\vv\times 1}$ is the bias term. 

As $f_\text{de}$ is a linear projection, the simplest situation is when $\mW$ is an orthogonal matrix and its inverse is equal to its transpose. Often, as the dimensionality of vector $\vz$ doesn't necessarily need to match that of word vectors $\vv$, $\mU$ is not a square matrix \footnote{As often the dimension of sentence vectors are equal to or large than that of word vectors, $\mW$ has more columns than rows. If it is not the case, then regulariser becomes $||\mW^\top\mW-\mI||_F$.}. To enforce invertibility on $\mW$, a row-wise orthonormal regularisation on $\mW$ is applied during learning, which leads to $\mW\mW^\top=\mI$, where $\mI$ is the identity matrix, thus the inverse function is simply $\vz = f_\text{de}^{-1}(\vx)=\mW^\top(\vx - \vb)$, which is easily computed. The regularisation formula is $||\mW\mW^\top-\mI||_F$, where $||\cdot||_F$ is the Frobenius norm. Specifically, the update rule \citep{Ciss2017ParsevalNI} for the regularisation is:
\begin{align}
\mW:=(1+\beta)\mW-\beta(\mW\mW^\top)\mW \nonumber
\end{align}
The usage of the decoder during training and testing is defined as follows:
\begin{align}
    \text{Training:} \hspace{0.1cm} & \vx = f_{\text{de}}(\vz)=\mW\vz + \vb \nonumber \\
    \text{Testing:}  \hspace{0.1cm} & \vz = f_\text{de}^{-1}(\vx)=\mW^\top(\vx - \vb) \nonumber 
\end{align}
Therefore, the decoder is also utilised after learning to serve as a linear encoder in addition to the RNN encoder.

\subsubsection{Bijective Functions}

A general case is to use a bijective function as the decoder, as the bijective functions are naturally invertible. However, the inverse of a bijective function could be hard to find and 
its calculation could also be computationally intense. 

A family of bijective transformation was designed in NICE \cite{Dinh2014NICENI}, and the simplest continuous bijective function $f:\sR^D\rightarrow\sR^D$ and its inverse $f^{-1}$ is defined as:
\begin{align}
h: \hspace{0.5cm} \vy_1 &= \vx_1, & \vy_2 &= \vx_2+m(\vx_1) \nonumber \\
h^{-1}: \hspace{0.5cm} \vx_1 &= \vy_1, & \vx_2 &= \vy_2-m(\vy_1) \nonumber
\end{align}
where $\vx_1$ is a $d$-dimensional partition of the input $\vx\in\sR^D$, and $m:\sR^d\rightarrow\sR^{D-d}$ is an arbitrary continuous function, which could be a trainable multi-layer feedforward neural network with non-linear activation functions. It is named as an `additive coupling layer' \cite{Dinh2014NICENI}, which has unit Jacobian determinant. To allow the learning system to explore more powerful transformation, we follow the design of the `affine coupling layer' \cite{Dinh2016DensityEU}:
\begin{align}
h: \hspace{0.2cm} \vy_1 &= \vx_1, & \vy_2 &= \vx_2 \odot \text{exp}(s(\vx_1)) + t(\vx_1) \nonumber \\
h^{-1}: \hspace{0.2cm} \vx_1 &= \vy_1, & \vx_2 &= (\vy_2-t(\vy_1)) \odot \text{exp}(-s(\vx_1))  \nonumber
\end{align}
where $s:\sR^d\rightarrow\sR^{D-d}$ and $t:\sR^d\rightarrow\sR^{D-d}$ are both neural networks with linear output units.

The requirement of the continuous bijective transformation is that, the dimensionality of the input $\vx$ and the output $\vy$ need to match exactly. In our case, the output $\vx\in\sR^{d_\vv}$ of the decoding function $f_{\text{de}}$ has lower dimensionality than the input $\vz\in\sR^{d_\vz}$ does. Our solution is to add an orthonormal regularised linear projection before the bijective function to transform the vector representation of a sentence to the desired dimension. 

The usage of the decoder that is composed of a bijective function and a regularised linear projection during training and testing is defined as:
\begin{align}
    \text{Training:} \hspace{0.1cm} & \vx = f_{\text{de}}(\vz) = h(\mW\vz + \vb) \nonumber \\
    \text{Testing:}  \hspace{0.1cm} & \vz = f_\text{de}^{-1}(\vx) = \mW^\top(h^{-1}(\vx) - \vb) \nonumber 
\end{align}

\subsection{Using Decoder in the Test Phase}

As the decoder is easily invertible, it is also used to produce vector representations. The post-processing step \cite{Arora2017ASB} that removes the top principal component is applied on the representations from $f_\text{en}$ and $f^{-1}_\text{de}$ individually. In the following sections, $\vz_\text{en}$ denotes the post-processed representation from $f_\text{en}$, and $\vz_\text{de}$ from $f^{-1}_\text{de}$. Since $f_\text{en}$ and $f^{-1}_\text{de}$ naturally process sentences in distinctive ways, it is reasonable to expect that the ensemble of $\vz_\text{en}$ and $\vz_\text{de}$ will outperform each of them.

\section{Experimental Design}
\label{sec:experiment}
Experiments are conducted in PyTorch \cite{paszke2017automatic}, with evaluation  using the SentEval package \cite{Conneau2017SupervisedLO} with modifications to include the post-processing step. Word vectors $\vv_{w_j}$ are initialised with FastText \cite{Bojanowski2017EnrichingWV}, and fixed during learning.

\subsection{Unlabelled Corpora}
Two unlabelled corpora, including \textbf{B}ookCorpus \cite{Zhu2015AligningBA} and \textbf{U}MBC News Corpus \cite{han2013umbc_ebiquity}, are used to train models with invertible decoders. These corpora are referred as \textbf{B}, and \textbf{U} in Table \ref{unsupervised} and \ref{supervised}. The UMBC News Corpus is roughly twice as large as the BookCorpus, and the details are shown in Table \ref{stats}. 

\begin{table}[th]
\fontsize{8.5}{10}\selectfont
\begin{center}
    \vskip 0.15in
    \tabulinesep =_1pt^2pt
    \begin{tabu}to \textwidth{@{} c | c @{}}
    \toprule
        Name & \# of sentences \\
        \midrule
        BookCorpus (\textbf{B}) & 74 million  \\
        UMBC News (\textbf{U}) & 134.5 million  \\
    \bottomrule
\end{tabu}
\end{center}
\caption{\textbf{Summary statistics} of the two corpora used. For simplicity, the two corpora are referred to as \textbf{B} and \textbf{U} in the following tables respectively.}
\label{stats}
\end{table}

\subsection{Evaluation Tasks}

\subsection{Unsupervised Evaluation}
The unsupervised tasks include five tasks from SemEval Semantic Textual Similarity (STS) in 2012-2016 \cite{Agirre2015SemEval2015T2,Agirre2014SemEval2014T1,Agirre2016SemEval2016T1,Agirre2012SemEval2012T6,Agirre2013SEM2S} and the SemEval2014 Semantic Relatedness task (SICK-R) \cite{Marelli2014ASC}. 

The cosine similarity between vector representations of two sentences determines the textual similarity of two sentences, and the performance is reported in Pearson's correlation score between human-annotated labels and the model predictions on each dataset.

\subsection{Supervised Evaluation}
It includes Semantic relatedness (SICK) \cite{Marelli2014ASC}, SemEval (STS-B) \cite{Cer2017SemEval2017T1}, paraphrase detection (MRPC) \cite{Dolan2004UnsupervisedCO}, question-type classification (TREC) \cite{Li2002LearningQC}, movie review sentiment (MR) \cite{Pang2005SeeingSE}, Stanford Sentiment Treebank (SST) \cite{Socher2013RecursiveDM}, customer product reviews (CR) \cite{Hu2004MiningAS}, subjectivity/objectivity classification (SUBJ) \cite{Pang2004ASE}, opinion polarity (MPQA) \cite{Wiebe2005AnnotatingEO}.

In these tasks, MR, CR, SST, SUBJ, MPQA and MRPC are binary classification tasks, TREC is a multi-class classification task. SICK and MRPC require the same feature engineering method \cite{Tai2015ImprovedSR} in order to compose a vector from vector representations of two sentences to indicate the difference between them.

\begin{table*}[!ht]
\fontsize{8.5}{10}\selectfont
\begin{center}
    \vskip 0.15in
    \tabulinesep =_1pt^2pt
    \begin{tabu}to \textwidth{@{}c | c | c | c | c | c @{}}
        \toprule
        & & Unsupervised tasks & \multicolumn{3}{c}{Supervised tasks} \\
        \cmidrule{3-6}
        \textbf{Toronto} & \multirow{3}{*}{Hrs} & Avg of STS tasks  & Avg of & Avg of Binary-CLS tasks & \multirow{2}{*}{MRPC} \\ 
        \textbf{BookCorpus} & & (STS12-16, SICK14) & SICK-R, STS-B & (MR, CR, SUBJ, MPQA, SST) &  \\ 
        \bottomrule
        \toprule
        \multicolumn{6}{c}{Generative Objective with \textbf{Invertible} Linear Projection}
        \\ 
        \midrule
        $\vz_\text{en}$ & \multirow{3}{*}{3}  & 64.9 & 82.2 & 86.2 & 75.1/83.4 \\
        $\vz_\text{de}$ &                    & 67.6 & 82.2 & 85.2 & 73.7/82.5  \\
        ensemble$(\vz_\text{en}, \vz_\text{de})$ &   & 70.2 & 83.1 & 87.0 & 76.5/83.7  \\
        \midrule
        \multicolumn{6}{c}{Generative Objective with Linear Projection} \\
        \midrule
        $\vz_\text{en}$ & \multirow{3}{*}{3} & 54.6 \textcolor{red}{($\downarrow$10.3)} & 79.5 \textcolor{red}{($\downarrow$2.7)} & 85.6 \textcolor{red}{($\downarrow$0.6)} & 75.1/82.5 \\
        $\vz_\text{de}$ &                    &  69.5 \textcolor{green}{($\uparrow$1.9)} & 82.4 \textcolor{green}{($\uparrow$0.2)} & 84.7 \textcolor{red}{($\downarrow$0.5)} & 74.3/82.0 \\
        ensemble$(\vz_\text{en}, \vz_\text{de})$ &  & 66.9 \textcolor{red}{($\downarrow$3.3)} & 82.8 \textcolor{red}{($\downarrow$0.3)} &  86.3 \textcolor{red}{($\downarrow$0.7)} & 76.2/83.5 \\
        \bottomrule
        \toprule
        \multicolumn{6}{c}{Generative Objective with Bijective Transformation + \textbf{Invertible} Linear Projection}
        \\ 
        \midrule
        $\vz_\text{en}$ & \multirow{3}{*}{3.3}  & 67.1 & 82.1 & 85.4 & 74.3/82.2 \\
        $\vz_\text{de}$ &                    & 67.6 & 82.1 & 85.0 & 74.6/82.4  \\
        ensemble$(\vz_\text{en}, \vz_\text{de})$ &   & 70.0 & 82.9 & 86.5 & 76.2/83.0  \\
        \midrule
        \multicolumn{6}{c}{Generative Objective with Bijective Transformation +  Linear Projection} \\
        \midrule
        $\vz_\text{en}$ & \multirow{3}{*}{3.3} & 63.4 \textcolor{red}{($\downarrow$3.7)} & 81.7 \textcolor{red}{($\downarrow$0.4)} & 85.2 \textcolor{red}{($\downarrow$0.2)} & 76.9/84.3 \\
        $\vz_\text{de}$ &                    &  67.8 \textcolor{green}{($\uparrow$0.2)} &  82.2 \textcolor{green}{($\uparrow$0.1)} & 84.1 \textcolor{red}{($\downarrow$0.9)} & 74.7/82.0 \\
        ensemble$(\vz_\text{en}, \vz_\text{de})$ &  & 69.4 \textcolor{red}{($\downarrow$0.6)} & 82.5 \textcolor{red}{($\downarrow$0.4)} &  86.1 \textcolor{red}{($\downarrow$0.4)} & 76.6/83.4 \\
        \bottomrule        
        
        \end{tabu}
    \vskip -0.1in

\end{center}
\caption{\textbf{The effect of the invertible constraint on the linear projection}. The arrow and its associated value of a representation is the relative performance gain or loss compared to its comparison partner with the invertible constraint. As shown, the invertible constraint does help improve each representation, an ensures the ensemble of two encoding functions gives better performance. Better view in colour. }
\label{UMBCmvsv}
\end{table*}

\subsection{Hyperparameter Tuning}
The hyperparameters are tuned on the averaged scores on STS14 of the model trained on \textbf{B}ookCorpus, thus it is marked with a $^\star$ in tables to indicate potential overfitting. 

The hyperparameter setting for our model is summarised as follows: the batch size $N=512$, the dimension of sentence vectors $d_\vz=2048$, the dimension of word vectors $d_{\vv_{w_j}}=300$, the number of negative samples $K=5$, and the initial learning rate is $5\times 10^{-4}$ which is kept fixed during learning. The Adam optimiser \cite{Kingma2014AdamAM} with gradient clipping \cite{Pascanu2013OnTD} is applied for stable learning. Each model in our experiment is only trained for one epoch on the given training corpus. 

$\beta$ in the invertible constraint of the linear projection is set to be $0.01$, and after learning, all $300$ eigenvalues are close to $1$. For the bijective transformation, in order to make sure that each output unit is influenced by all input units, we stack four affine coupling layers in the bijective transformation \cite{Dinh2014NICENI}. The non-linear mappings $s$ and $t$ are both neural networks with one hidden layer with the rectified linear activation function.

\begin{table*}[!ht]
\fontsize{8.5}{10}\selectfont
\begin{center}
\vskip 0.15in
\tabulinesep =_1pt^2pt
\begin{tabu}to \textwidth{@{}  c || c | c | c | c || c c || c c || c | c  @{}}
\toprule
\multicolumn{11}{c}{$^1$\citet{Arora2017ASB};$^2$\citet{Wieting2015FromPD};$^3$\citet{Wieting2018Para};$^4$\citet{Conneau2017SupervisedLO};} \\
\multicolumn{11}{c}{$^5$\citet{Wieting2018Para};$^{6-10}$\citet{Agirre2012SemEval2012T6,Agirre2013SEM2S,Agirre2014SemEval2014T1,Agirre2015SemEval2015T2,Agirre2016SemEval2016T1};} \\
\multicolumn{11}{c}{$^{11}$\citet{Marelli2014ASC};$^{12}$\citet{Mikolov2017AdvancesIP}} \\
\midrule
\multirow{3}{*}{Task} & \multicolumn{6}{c||}{Un. Training} & \multicolumn{2}{c||}{Semi.} & \multicolumn{2}{c}{Su.} \\
\cmidrule{2-11}
 & \multicolumn{2}{c|}{\textbf{Linear}} & \multicolumn{2}{c|}{\textbf{Bijective}} & \multicolumn{2}{c||}{fastText} & \multicolumn{2}{c||}{$^2$PSL} & $^4$Infer & $^3$ParaNMT \\
\cmidrule{2-9}
 & \textbf{B} & \textbf{U} & \textbf{B} & \textbf{U} & $^{12}$avg & $^1$WR & $^1$avg & $^1$WR & Sent & (concat.) \\
 \bottomrule
$^6$STS12       & 61.5 & 61.3 & 60.8 & \textbf{62.7} &  58.3 & 58.8 & 52.8 & 59.5 & 58.2 & \underline{67.7} \\
$^7$STS13       & 61.3 & 61.8 & 60.7 & \textbf{62.2} &  51.0 & 59.9 & 46.4 & 61.8 & 48.5 & \underline{62.8} \\
$^8$STS14       & 71.6 & 72.1 & 72.1 & \textbf{73.2} &  65.2 & 69.4 & 59.5 & 73.5 & 67.1 & \underline{76.9} \\
$^9$STS15       & 76.1 & 76.9 & 76.6 & \textbf{77.6} &  67.7 & 74.2 & 60.0 & 76.3 & 71.1 & \underline{79.8} \\
$^{10}$STS16    & 74.8 & 76.1 & 75.8 & \textbf{76.9} &  64.3 & 72.4 & - & - & 71.2 & \underline{76.8} \\
$^{11}$SICK14   & \textbf{76.1} & 73.6 & 74.2 & 73.9 &  69.8 & 72.3  & 66.4 & 72.9 & 73.4 & - \\
\midrule
Average & 70.2 & 70.3 & 70.0 & \textbf{71.1} & 62.7 & 67.8 & - & - & 64.9 & - \\
\bottomrule
\end{tabu}
\end{center}
\caption{\textbf{Results on unsupervised evaluation tasks} (Pearson's $r\times 100$) . \textbf{Bold} numbers are the best results among unsupervised transfer models, and \underline{underlined} numbers are the best ones among all models. `WR' refers to the post-processing step that removes the top principal component.}
\label{unsupervised}
\end{table*}

\subsection{Representation Pooling}
Various pooling functions are applied to produce vector representations for input sentences.

For unsupervised evaluation tasks, as recommended in previous studies \cite{Pennington2014GloveGV,Kenter2016SiameseCO,Wieting2017RevisitingRN}, a global mean-pooling function is applied on both the output of the RNN encoder $f_\text{en}$ to produce a vector representation $\vz_\text{en}$ and the inverse of the decoder $f_\text{de}^{-1}$ to produce $\vz_\text{de}$.

For supervised evaluation tasks, three pooling functions, including global max-, min-, and mean-pooling, are applied on top of the encoder and the outputs from three pooling functions are concatenated  to serve as a vector representation for a given sentence. The same representation pooling strategy is applied on the inverse of the decoder. 

The reason for applying different representation pooling strategies for two categories of tasks is:

\textbf{(1)} cosine similarity of two vector representations is directly calculated in unsupervised evaluation tasks to determine the textual similarity of two sentences, and it suffers from the curse-of-dimensionality  \cite{Donoho2000AideMemoireH}, which leads  to more equidistantly distributed  representations for higher  dimensional vector representations decreasing the difference among similarity scores.

\textbf{(2)} given  Cover's theorem \cite{cover1965geometrical} and the blessings-of-dimensionality property, it is more likely for the data points to be linearly separable when they are presented in high dimensional space, and in the supervised evaluation tasks, high dimensional vector representations are preferred as a linear classifier will be learnt to evaluate how likely the produced sentence representations are linearly separable; 

\textbf{(3)} in our case, both the encoder and the inverse of the decoder are capable of producing a vector representation per time step in a given sentence, although during training, only the last one is regarded as the sentence representation for the fast training speed, it is more reasonable to make use of all representations at all time steps with various pooling functions to compute a vector representations to produce high-quality sentence representations that excel the downstream tasks.

\section{Discussion}
\label{sec:discussion}
It is worth discussing the motivation of the model design and the observations in our experiments. As mentioned as one of the take-away messages \cite{wieting2018no}, to demonstrate the effectiveness of the invertible constraint, the comparison of our model with the constraint and its own variants use the same word embeddings from FastText \cite{Bojanowski2017EnrichingWV} and have the same dimensionaility of sentence representations during learning, and use the same classifier on top of the produced representations with the same hyperparameter settings.

Overall, given the performance of the inverse of each decoder presented in Table \ref{unsupervised} and \ref{supervised}, it is reasonable to state that the inverse of the decoder provides high-quality sentence representations as well as the encoder does. However, there is no significant difference between the two decoders in terms of the performance on the downstream tasks. In this section, observations and thoughts are presented based on the analyses of our model with the invertible constraint.

\subsection{Effect of Invertible Constraint}
The motivation of enforcing the invertible constraint on the decoder during learning is to make it usable and potentially helpful during testing in terms of boosting the performance of the lone RNN encoder in the encoder-decoder models (instead of ignoring the decoder part after learning). Therefore, it is important to check the necessity of the invertible constraint on the decoders. 

A model with the same hyperparameter settings but without the invertible constraint is trained as the baseline model, and  macro-averaged results that summarise the same type of tasks are presented in Table \ref{UMBCmvsv}. 

As noted in the prior work \cite{Hill2016LearningDR},  there exists significant inconsistency between the group of unsupervised tasks and the group of supervised ones, it is possible for a model to excel on one group of tasks but fail on the other one. As presented in our table, the inverse of the decoder tends to perform better than the encoder on unsupervised tasks, and the situation reverses when it comes to the supervised ones. 

In our model, the invertible constraint \textbf{helps the RNN encoder} $f_\text{en}$ to perform better on the unsupervised evaluation tasks, and \textbf{helps the inverse of the decoder} $f_\text{de}^{-1}$ to provide better results on single sentence classification tasks. An interesting observation is that, by enforcing the invertible constraint, the model learns to sacrifice the performance of $f_\text{de}^{-1}$ and improve the performance of $f_\text{en}$ on unsupervised tasks to mitigate the gap between the two encoding functions, which leads to more aligned vector representations between $f_\text{en}$ and $f_\text{de}^{-1}$.

\subsection{Effect on Ensemble}

Although encouraging the invertible constraint leads to slightly poorer performance of $f_\text{de}^{-1}$ on unsupervised tasks, it generally leads to better sentence representations when the ensemble of the encoder $f_\text{en}$ and the inverse of the decoder $f_\text{de}^{-1}$ is considered. Specifically, for unsupervised tasks, the ensemble is an average of two vector representations produced from two encoding functions during the testing time, and for supervised tasks, the concatenation of two representations is regarded as the representation of a given sentence. The ensemble method is recommended in prior work \cite{Pennington2014GloveGV,Levy2015ImprovingDS,Wieting2017RevisitingRN,McCann2017LearnedIT,Tang2018SpeedingUC,wieting2018no}.

As presented in Table \ref{UMBCmvsv}, on unsupervised evaluation tasks (STS12-16 and SICK14), the ensemble of two encoding functions is averaging, which benefits from aligning representations from $f_\text{en}$ and $f_\text{de}^{-1}$ by enforcing the invertible constraint. While in the learning system without the invertible constraint, the ensemble of two encoding functions provides worse performance than $f_\text{de}^{-1}$.

On supervised evaluation tasks, as the ensemble method is concatenation and a linear model is applied on top of the concatenated representations, as long as the two encoding functions process sentences distinctively, the linear classifier is capable of picking relevant feature dimensions from both encoding functions to make good predictions, thus there is no significant difference between our model with and without invertible constraint.

\subsection{Effect of Learning}
Recent research \cite{wieting2018no} showed that the improvement on the supervised evaluation tasks led by learning from labelled or unlabelled corpora is rather insignificant compared to random initialised projections on top of pretrained word vectors. Another interesting direction of research that utilises probabilistic random walk models on the unit sphere \cite{Arora2016ALV,Arora2017ASB,Ethayarajh2018UnsupervisedRW} derived several simple yet effective post-processing methods that operate on pretrained word vectors and are able to boost the performance of the averaged word vectors as the sentence representation on unsupervised tasks. While these papers reveal interesting aspects of the downstream tasks and question the need for optimising a learning objective, our results show that learning on unlabelled corpora helps.

On \textbf{unsupervised} evaluation tasks, in order to show that learning from an unlabelled corpus helps, the performance of our learnt representations should be directly compared with the pretrained word vectors, FastText in our system, at the same dimensionality with the same post-processing \cite{Arora2017ASB}. The word vectors are scattered in the 300-dimensional space, and our model has a decoder that is learnt to project a sentence representation $\vz\in\sR^{d_\vz}$ to $\vx=f_\text{de}(\vz;\boldsymbol{\phi})\in\sR^{300}$. The results of our learnt representations and averaged word vectors with the same postprocessing  are presented in Table \ref{D300}. 

As shown in the Table \ref{D300}, the performance of our learnt system is better than FastText at the same dimensionality. It is worth mentioning that, in our system, the final representation is an average of postprocessed word vectors and the learnt representations $\vx$, and the invertible constraint guarantees that the ensemble of both gives better performance. Otherwise, as discussed in the previous section, an ensemble of postprocessed word vectors and some random encoders won't necessarily lead to stronger results. Table \ref{unsupervised} also provides evidence for the effectiveness of learning on the unsupervised evaluation tasks.

\begin{table}[ht]
\fontsize{8.5}{10}\selectfont
\begin{center}

\vskip 0.15in
\tabulinesep =_1pt^2pt
\begin{tabu}to \textwidth{@{} c | c c c @{}}

\toprule
Task & Linear & Bijective & FastText+WR \\
\midrule
STS12  & 60.7 & \textbf{60.9} & 58.8 \\
STS13  & \textbf{61.1} & 60.0 & 59.9 \\
STS14  & \textbf{71.7} & \textbf{71.7} & 69.4 \\
STS15  & \textbf{75.9} & 75.4 & 74.2 \\
STS16  & \textbf{74.9} & 73.5 & 72.4 \\
SICK14 & 75.7 & \textbf{75.8} & 72.3 \\
\midrule
Average & \textbf{70.0} & 69.6 & 67.8 \\
\bottomrule
\end{tabu}
\end{center}
\caption{\textbf{Comparison} of the learnt representations in our system with the \textbf{same dimensionality} as pretrained word vectors on unsupervised evaluation tasks. The encoding function that is learnt to compose a sentence representation from pretrained word vectors outperforms averaging word vectors, which supports our argument that learning helps to produce higher-quality sentence representations.}
\label{D300}
\end{table}

\begin{table*}[t]
\fontsize{8.5}{10}\selectfont
\begin{center}

\vskip 0.15in
\tabulinesep =_2pt^2pt
\begin{tabu}to \textwidth{@{}l| c | c c c | c c c c c c@{}}
\toprule
\multicolumn{11}{c}{$^1$\citet{Conneau2017SupervisedLO};$^2$\citet{Hill2016LearningDR}; $^3$\citet{Kiros2015SkipThoughtV};$^4$\citet{Ba2016LayerN};$^5$\citet{Gan2017LearningGS};} \\
\multicolumn{11}{c}{$^6$\citet{Jernite2017DiscourseBasedOF};$^7$\citet{Nie2017DisSentSR};$^8$\citet{Zhao2015SelfAdaptiveHS};$^9$\citet{logeswaran2018an};$^{10}$\citet{Marelli2014ASC};}\\
\multicolumn{11}{c}{$^{11}$\citet{Dolan2004UnsupervisedCO};$^{12}$\citet{Li2002LearningQC};$^{13}$\citet{Pang2005SeeingSE};$^{14}$\citet{Hu2004MiningAS}}\\
\multicolumn{11}{c}{$^{15}$\citet{Pang2004ASE};$^{16}$\citet{Wiebe2005AnnotatingEO};$^{17}$\citet{Socher2013RecursiveDM};$^{18}$\citet{wieting2018no}}\\


\bottomrule
Model & Hrs & $^{10}$SICK-R & $^{10}$SICK-E & $^{11}$MRPC & $^{12}$TREC & $^{13}$MR & $^{14}$CR & $^{15}$SUBJ & $^{16}$MPQA & $^{17}$SST \\
\toprule
\multicolumn{11}{c}{\textbf{Supervised task-dependent training - No transfer learning}} \\
\midrule
$^8$AdaSent & - & - & - & - & 92.4 & 83.1 & \underline{86.3} & \underline{95.5} & \underline{93.3} & - \\
$^1$TF-KLD & - & - & - & \underline{80.4}/\underline{85.9} & - & - & - & - & - & - \\
\bottomrule
\toprule
\multicolumn{11}{c}{\textbf{Supervised training - Transfer learning}} \\
\midrule
$^1$InferSent & $<$24 & \underline{88.4} & \underline{86.3} & 76.2/83.1 & 88.2 & 81.1 & \underline{86.3} & 92.4 & 90.2 & 84.6 \\
\bottomrule
\toprule
\multicolumn{11}{c}{\textbf{Unsupervised training with ordered sentences}} \\
\midrule
$^2$FastSent+AE  & 2   & -      & -    & 71.2/79.1 & 80.4 & 71.8 & 76.5 & 88.8 & 81.5 & - \\
$^4$ST+LN        & 720 & 85.8 & 79.5 & -         & 88.4 & 79.4 & 83.1 & 93.7 & 89.3 & 82.9 \\
$^5$CNN-LSTM & -   & 86.2 & -    & 76.5/83.8 & 92.6 & 77.8 & 82.1 & 93.6 & 89.4 & - \\
\midrule
$^6$DiscSent & 8  & -    & -    & 75.0/ -   & 87.2 & -    & -    & 93.0 & -    & - \\
$^7$DisSent & -  & 79.1 & 80.3 & - / -     & 84.6 & 82.5 & 80.2 & 92.4 & 89.6 & 82.9 \\
$^9$MC-QT      & 11 & 86.8 & -    & 76.9/\textbf{84.0} & \underline{\textbf{92.8}} & 80.4 & \textbf{85.2} & 93.9 & 89.4 & -\\
\midrule

\textbf{B} - Bijective $\vz$        & 3.3 & 87.9 & 84.5 & 76.2/83.0 & 89.6 & 80.3 & 82.6 & 94.6 & 89.3 & 85.6 \\
\textbf{B} - Linear $\vz$           & 3 & \textbf{88.1} & 85.2 & 76.5/83.7 & 90.0 & \textbf{81.3} & 83.5 & 94.6 & \textbf{89.5} & \textbf{85.9} \\
\midrule

\textbf{U} - Bijective $\vz$ & 10 & 87.8 & 85.2 & 76.4/83.7 & 90.8 & 80.9 & 82.7 & 94.6 & 89.2 & 83.3 \\
\textbf{U} - Linear $\vz$    & 8.8 & 87.8 & \textbf{85.9} & \textbf{77.5}/83.8 & 92.2 & \textbf{81.3} & 83.4 & \textbf{94.7} & \textbf{89.5} & \textbf{85.9} \\
\bottomrule
\toprule
\multicolumn{11}{c}{\textbf{No training - $^{18}$Global max-pooling on top of random projection}} \\
\midrule
BOREP    & 0 & 85.9 & 84.3 & 73.7/ - & 89.5 & 78.6 & 79.9 & 93.0 & 88.8 & 82.5 \\
RandLSTM & 0 & 86.6 & 83.0 & 74.7/ - & 88.4 & 78.2 & 79.9 & 92.8 & 88.2 & 83.2 \\
ESN      & 0 & 87.2 & 85.1 & 75.3/ - & 92.2 & 79.1 & 80.2 & 93.4 & 88.9 & 84.6 \\
\bottomrule
\end{tabu}
\vskip -0.1in
\end{center}
\caption{\textbf{Results on supervised evaluation tasks.} \textbf{Bold} numbers are the best results among unsupervised transfer models with ordered sentences, and \underline{underlined} numbers are the best ones among all models.}

\label{supervised}
\end{table*}

On \textbf{supervised} evaluation tasks, we agree that higher  dimensional vector representations give better results on the downstream tasks. Compared to random projections with $4096\times 6$ output dimensions, learning from unlabelled corpora
leverages the distributional similarity \cite{firth57synopsis} at the sentence-level into the learnt representations and potentially helps capture the meaning of a sentence. In our system, the raw representations are in 2400-dimensional space, and the use of various pooling functions expands it to $2048\times 6$ dimensions, which is half as large as the random projection dimension and still yields better performance. Both our models and random projections with no training are presented in Table \ref{supervised}.

The evidence from both sets of downstream tasks support our argument that learning from unlabelled corpora helps the representations capture meaning of sentences. However, current ways of incorporating the distributional hypothesis only utilise it as a weak and noisy supervision, which might limit the quality of the learnt sentence representations.

\section{Conclusion}

Two types of decoders, including an orthonormal regularised linear projection and a bijective transformation, whose inverses can be derived effortlessly, are presented in order to utilise the decoder as another encoder in the testing phase. The experiments and comparisons are conducted on two large unlabelled corpora, and the performance on the downstream tasks shows the high usability and generalisation ability of the decoders in testing.

Analyses show that the invertible constraint enforced on the decoder encourages each one to learn from the other one during learning, and provides improved encoding functions after learning. Ensemble of the encoder and the inverse of the decoder gives even better performance when the invertible constraint is applied on the decoder side. Furthermore, by comparing with prior work, we argue that learning from unlabelled corpora indeed helps to improve the sentence representations, although the current way of utilising corpora might not be optimal.

We view this as unifying the generative and discriminative objectives for unsupervised sentence representation learning, as it is trained with a generative objective which when inverted can be seen as creating a discriminative target.   

Our proposed method in our implementation doesn't provide extremely good performance on the downstream tasks, but we see our method as an opportunity to fuse all possible components in a model, even a usually discarded decoder, to produce sentence representations. Future work could potentially expand our work into end-to-end invertible model that is able to produce high-quality representations by omnidirectional computations.

\section*{Acknowledgements}
Many Thanks to Andrew Ying for helpful clarifications on several concepts.

\bibliography{acl2019}

\begin{thebibliography}{61}
\expandafter\ifx\csname natexlab\endcsname\relax\def\natexlab#1{#1}\fi

\bibitem[{Agirre et~al.(2015)Agirre, Banea, Cardie, Cer, Diab, Gonzalez-Agirre,
  Guo, Lopez-Gazpio, Maritxalar, Mihalcea, Rigau, Uria, and
  Wiebe}]{Agirre2015SemEval2015T2}
Eneko Agirre, Carmen Banea, Claire Cardie, Daniel~M. Cer, Mona~T. Diab, Aitor
  Gonzalez-Agirre, Weiwei Guo, I{\~n}igo Lopez-Gazpio, Montse Maritxalar, Rada
  Mihalcea, German Rigau, Larraitz Uria, and Janyce Wiebe. 2015.
\newblock Semeval-2015 task 2: Semantic textual similarity, english, spanish
  and pilot on interpretability.
\newblock In \emph{SemEval@NAACL-HLT}.

\bibitem[{Agirre et~al.(2014)Agirre, Banea, Cardie, Cer, Diab, Gonzalez-Agirre,
  Guo, Mihalcea, Rigau, and Wiebe}]{Agirre2014SemEval2014T1}
Eneko Agirre, Carmen Banea, Claire Cardie, Daniel~M. Cer, Mona~T. Diab, Aitor
  Gonzalez-Agirre, Weiwei Guo, Rada Mihalcea, German Rigau, and Janyce Wiebe.
  2014.
\newblock Semeval-2014 task 10: Multilingual semantic textual similarity.
\newblock In \emph{SemEval@COLING}.

\bibitem[{Agirre et~al.(2016)Agirre, Banea, Cer, Diab, Gonzalez-Agirre,
  Mihalcea, Rigau, and Wiebe}]{Agirre2016SemEval2016T1}
Eneko Agirre, Carmen Banea, Daniel~M. Cer, Mona~T. Diab, Aitor Gonzalez-Agirre,
  Rada Mihalcea, German Rigau, and Janyce Wiebe. 2016.
\newblock Semeval-2016 task 1: Semantic textual similarity, monolingual and
  cross-lingual evaluation.
\newblock In \emph{SemEval@NAACL-HLT}.

\bibitem[{Agirre et~al.(2012)Agirre, Cer, Diab, and
  Gonzalez-Agirre}]{Agirre2012SemEval2012T6}
Eneko Agirre, Daniel~M. Cer, Mona~T. Diab, and Aitor Gonzalez-Agirre. 2012.
\newblock Semeval-2012 task 6: A pilot on semantic textual similarity.
\newblock In \emph{SemEval@NAACL-HLT}.

\bibitem[{Agirre et~al.(2013)Agirre, Cer, Diab, Gonzalez-Agirre, and
  Guo}]{Agirre2013SEM2S}
Eneko Agirre, Daniel~M. Cer, Mona~T. Diab, Aitor Gonzalez-Agirre, and Weiwei
  Guo. 2013.
\newblock *sem 2013 shared task: Semantic textual similarity.
\newblock In \emph{*SEM@NAACL-HLT}.

\bibitem[{Arora et~al.(2016)Arora, Li, Liang, Ma, and Risteski}]{Arora2016ALV}
Sanjeev Arora, Yuanzhi Li, Yingyu Liang, Tengyu Ma, and Andrej Risteski. 2016.
\newblock A latent variable model approach to pmi-based word embeddings.
\newblock \emph{TACL}, 4:385--399.

\bibitem[{Arora et~al.(2017)Arora, Liang, and Ma}]{Arora2017ASB}
Sanjeev Arora, Yingyu Liang, and Tengyu Ma. 2017.
\newblock A simple but tough-to-beat baseline for sentence embeddings.
\newblock In \emph{International Conference on Learning Representations}.

\bibitem[{Ba et~al.(2016)Ba, Kiros, and Hinton}]{Ba2016LayerN}
Jimmy Ba, Ryan Kiros, and Geoffrey~E. Hinton. 2016.
\newblock Layer normalization.
\newblock \emph{CoRR}, abs/1607.06450.

\bibitem[{Bojanowski et~al.(2017)Bojanowski, Grave, Joulin, and
  Mikolov}]{Bojanowski2017EnrichingWV}
Piotr Bojanowski, Edouard Grave, Armand Joulin, and Tomas Mikolov. 2017.
\newblock Enriching word vectors with subword information.
\newblock \emph{TACL}, 5:135--146.

\bibitem[{Cer et~al.(2017)Cer, Diab, Agirre, Lopez-Gazpio, and
  Specia}]{Cer2017SemEval2017T1}
Daniel~M. Cer, Mona~T. Diab, Eneko Agirre, I{\~n}igo Lopez-Gazpio, and Lucia
  Specia. 2017.
\newblock Semeval-2017 task 1: Semantic textual similarity multilingual and
  crosslingual focused evaluation.
\newblock In \emph{SemEval@ACL}.

\bibitem[{Chung et~al.(2014)Chung, Gulcehre, Cho, and
  Bengio}]{Chung2014EmpiricalEO}
Junyoung Chung, Caglar Gulcehre, KyungHyun Cho, and Yoshua Bengio. 2014.
\newblock Empirical evaluation of gated recurrent neural networks on sequence
  modeling.
\newblock \emph{arXiv preprint arXiv:1412.3555}.

\bibitem[{Ciss{\'e} et~al.(2017)Ciss{\'e}, Bojanowski, Grave, Dauphin, and
  Usunier}]{Ciss2017ParsevalNI}
Moustapha Ciss{\'e}, Piotr Bojanowski, Edouard Grave, Yann Dauphin, and Nicolas
  Usunier. 2017.
\newblock Parseval networks: Improving robustness to adversarial examples.
\newblock In \emph{ICML}.

\bibitem[{Conneau et~al.(2017)Conneau, Kiela, Schwenk, Barrault, and
  Bordes}]{Conneau2017SupervisedLO}
Alexis Conneau, Douwe Kiela, Holger Schwenk, Lo\"{i}c Barrault, and Antoine
  Bordes. 2017.
\newblock Supervised learning of universal sentence representations from
  natural language inference data.
\newblock In \emph{EMNLP}.

\bibitem[{Cover(1965)}]{cover1965geometrical}
Thomas~M Cover. 1965.
\newblock Geometrical and statistical properties of systems of linear
  inequalities with applications in pattern recognition.
\newblock \emph{IEEE transactions on electronic computers}, (3):326--334.

\bibitem[{Devlin et~al.(2018)Devlin, Chang, Lee, and
  Toutanova}]{Devlin2018BERTPO}
Jacob Devlin, Ming-Wei Chang, Kenton Lee, and Kristina Toutanova. 2018.
\newblock Bert: Pre-training of deep bidirectional transformers for language
  understanding.
\newblock \emph{CoRR}, abs/1810.04805.

\bibitem[{Dinh et~al.(2014)Dinh, Krueger, and Bengio}]{Dinh2014NICENI}
Laurent Dinh, David Krueger, and Yoshua Bengio. 2014.
\newblock Nice: Non-linear independent components estimation.
\newblock \emph{CoRR}, abs/1410.8516.

\bibitem[{Dinh et~al.(2016)Dinh, Sohl-Dickstein, and
  Bengio}]{Dinh2016DensityEU}
Laurent Dinh, Jascha Sohl-Dickstein, and Samy Bengio. 2016.
\newblock Density estimation using real nvp.
\newblock \emph{CoRR}, abs/1605.08803.

\bibitem[{Dolan et~al.(2004)Dolan, Quirk, and
  Brockett}]{Dolan2004UnsupervisedCO}
William~B. Dolan, Chris Quirk, and Chris Brockett. 2004.
\newblock Unsupervised construction of large paraphrase corpora: Exploiting
  massively parallel news sources.
\newblock In \emph{COLING}.

\bibitem[{Donoho(2000)}]{Donoho2000AideMemoireH}
David~L. Donoho. 2000.
\newblock Aide-memoire . high-dimensional data analysis : The curses and
  blessings of dimensionality.

\bibitem[{Ethayarajh(2018)}]{Ethayarajh2018UnsupervisedRW}
Kawin Ethayarajh. 2018.
\newblock Unsupervised random walk sentence embeddings: A strong but simple
  baseline.
\newblock In \emph{Rep4NLP@ACL}.

\bibitem[{Firth(1957)}]{firth57synopsis}
J.~R. Firth. 1957.
\newblock A synopsis of linguistic theory.

\bibitem[{Gan et~al.(2017)Gan, Pu, Henao, Li, He, and
  Carin}]{Gan2017LearningGS}
Zhe Gan, Yunchen Pu, Ricardo Henao, Chunyuan Li, Xiaodong He, and Lawrence
  Carin. 2017.
\newblock Learning generic sentence representations using convolutional neural
  networks.
\newblock In \emph{EMNLP}.

\bibitem[{Han et~al.(2013)Han, Kashyap, Finin, Mayfield, and
  Weese}]{han2013umbc_ebiquity}
Lushan Han, Abhay~L Kashyap, Tim Finin, James Mayfield, and Jonathan Weese.
  2013.
\newblock Umbc\_ebiquity-core: semantic textual similarity systems.
\newblock In \emph{Second Joint Conference on Lexical and Computational
  Semantics (* SEM), Volume 1: Proceedings of the Main Conference and the
  Shared Task: Semantic Textual Similarity}, volume~1, pages 44--52.

\bibitem[{Harris(1954)}]{harris1954distributional}
Zellig~S Harris. 1954.
\newblock Distributional structure.
\newblock \emph{Word}, 10(2-3):146--162.

\bibitem[{Hill et~al.(2016)Hill, Cho, and Korhonen}]{Hill2016LearningDR}
Felix Hill, Kyunghyun Cho, and Anna Korhonen. 2016.
\newblock Learning distributed representations of sentences from unlabelled
  data.
\newblock In \emph{HLT-NAACL}.

\bibitem[{Hu and Liu(2004)}]{Hu2004MiningAS}
Minqing Hu and Bing Liu. 2004.
\newblock Mining and summarizing customer reviews.
\newblock In \emph{KDD}.

\bibitem[{Jernite et~al.(2017)Jernite, Bowman, and
  Sontag}]{Jernite2017DiscourseBasedOF}
Yacine Jernite, Samuel~R. Bowman, and David Sontag. 2017.
\newblock Discourse-based objectives for fast unsupervised sentence
  representation learning.
\newblock \emph{CoRR}, abs/1705.00557.

\bibitem[{Kenter et~al.(2016)Kenter, Borisov, and
  de~Rijke}]{Kenter2016SiameseCO}
Tom Kenter, Alexey Borisov, and Maarten de~Rijke. 2016.
\newblock Siamese cbow: Optimizing word embeddings for sentence
  representations.
\newblock In \emph{ACL}.

\bibitem[{Kingma and Ba(2014)}]{Kingma2014AdamAM}
Diederik Kingma and Jimmy Ba. 2014.
\newblock Adam: A method for stochastic optimization.
\newblock \emph{arXiv preprint arXiv:1412.6980}.

\bibitem[{Kingma et~al.(2016)Kingma, Salimans, and
  Welling}]{Kingma2016ImprovingVI}
Diederik~P. Kingma, Tim Salimans, and Max Welling. 2016.
\newblock Improving variational inference with inverse autoregressive flow.
\newblock \emph{CoRR}, abs/1606.04934.

\bibitem[{Kiros et~al.(2015)Kiros, Zhu, Salakhutdinov, Zemel, Urtasun,
  Torralba, and Fidler}]{Kiros2015SkipThoughtV}
Jamie~Ryan Kiros, Yukun Zhu, Ruslan Salakhutdinov, Richard~S. Zemel, Raquel
  Urtasun, Antonio Torralba, and Sanja Fidler. 2015.
\newblock Skip-thought vectors.
\newblock In \emph{NIPS}.

\bibitem[{Lebret and Collobert(2014)}]{Lebret2014WordET}
R{\'e}mi Lebret and Ronan Collobert. 2014.
\newblock Word embeddings through hellinger pca.
\newblock In \emph{EACL}.

\bibitem[{Levy et~al.(2015)Levy, Goldberg, and Dagan}]{Levy2015ImprovingDS}
Omer Levy, Yoav Goldberg, and Ido Dagan. 2015.
\newblock Improving distributional similarity with lessons learned from word
  embeddings.
\newblock \emph{TACL}, 3:211--225.

\bibitem[{Li and Hovy(2014)}]{Li2014AMO}
Jiwei Li and Eduard~H. Hovy. 2014.
\newblock A model of coherence based on distributed sentence representation.
\newblock In \emph{EMNLP}.

\bibitem[{Li and Roth(2002)}]{Li2002LearningQC}
Xin Li and Dan Roth. 2002.
\newblock Learning question classifiers.
\newblock In \emph{COLING}.

\bibitem[{Logeswaran and Lee(2018)}]{logeswaran2018an}
Lajanugen Logeswaran and Honglak Lee. 2018.
\newblock An efficient framework for learning sentence representations.
\newblock In \emph{ICLR}.

\bibitem[{Marelli et~al.(2014)Marelli, Menini, Baroni, Bentivogli, Bernardi,
  and Zamparelli}]{Marelli2014ASC}
Marco Marelli, Stefano Menini, Marco Baroni, Luisa Bentivogli, Raffaella
  Bernardi, and Roberto Zamparelli. 2014.
\newblock A sick cure for the evaluation of compositional distributional
  semantic models.
\newblock In \emph{LREC}.

\bibitem[{McCann et~al.(2017)McCann, Bradbury, Xiong, and
  Socher}]{McCann2017LearnedIT}
Bryan McCann, James Bradbury, Caiming Xiong, and Richard Socher. 2017.
\newblock Learned in translation: Contextualized word vectors.
\newblock In \emph{NIPS}.

\bibitem[{Mikolov et~al.(2013{\natexlab{a}})Mikolov, Chen, Corrado, and
  Dean}]{Mikolov2013EfficientEO}
Tomas Mikolov, Kai Chen, Greg Corrado, and Jeffrey Dean. 2013{\natexlab{a}}.
\newblock Efficient estimation of word representations in vector space.
\newblock \emph{arXiv preprint arXiv:1301.3781}.

\bibitem[{Mikolov et~al.(2017)Mikolov, Grave, Bojanowski, Puhrsch, and
  Joulin}]{Mikolov2017AdvancesIP}
Tomas Mikolov, Edouard Grave, Piotr Bojanowski, Christian Puhrsch, and Armand
  Joulin. 2017.
\newblock Advances in pre-training distributed word representations.
\newblock \emph{CoRR}, abs/1712.09405.

\bibitem[{Mikolov et~al.(2013{\natexlab{b}})Mikolov, Sutskever, Chen, Corrado,
  and Dean}]{Mikolov2013DistributedRO}
Tomas Mikolov, Ilya Sutskever, Kai Chen, Gregory~S. Corrado, and Jeffrey Dean.
  2013{\natexlab{b}}.
\newblock Distributed representations of words and phrases and their
  compositionality.
\newblock In \emph{NIPS}.

\bibitem[{Nie et~al.(2017)Nie, Bennett, and Goodman}]{Nie2017DisSentSR}
Allen Nie, Erin~D. Bennett, and Noah~D. Goodman. 2017.
\newblock Dissent: Sentence representation learning from explicit discourse
  relations.
\newblock \emph{CoRR}, abs/1710.04334.

\bibitem[{Pang and Lee(2004)}]{Pang2004ASE}
Bo~Pang and Lillian Lee. 2004.
\newblock A sentimental education: Sentiment analysis using subjectivity
  summarization based on minimum cuts.
\newblock In \emph{ACL}.

\bibitem[{Pang and Lee(2005)}]{Pang2005SeeingSE}
Bo~Pang and Lillian Lee. 2005.
\newblock Seeing stars: Exploiting class relationships for sentiment
  categorization with respect to rating scales.
\newblock In \emph{ACL}.

\bibitem[{Pascanu et~al.(2013)Pascanu, Mikolov, and Bengio}]{Pascanu2013OnTD}
Razvan Pascanu, Tomas Mikolov, and Yoshua Bengio. 2013.
\newblock On the difficulty of training recurrent neural networks.
\newblock In \emph{ICML}.

\bibitem[{Paszke et~al.(2017)Paszke, Gross, Chintala, Chanan, Yang, DeVito,
  Lin, Desmaison, Antiga, and Lerer}]{paszke2017automatic}
Adam Paszke, Sam Gross, Soumith Chintala, Gregory Chanan, Edward Yang, Zachary
  DeVito, Zeming Lin, Alban Desmaison, Luca Antiga, and Adam Lerer. 2017.
\newblock Automatic differentiation in pytorch.
\newblock In \emph{NIPS-W}.

\bibitem[{Pennington et~al.(2014)Pennington, Socher, and
  Manning}]{Pennington2014GloveGV}
Jeffrey Pennington, Richard Socher, and Christopher~D. Manning. 2014.
\newblock Glove: Global vectors for word representation.
\newblock In \emph{EMNLP}.

\bibitem[{Peters et~al.(2018)Peters, Neumann, Iyyer, Gardner, Clark, Lee, and
  Zettlemoyer}]{Peters2018DeepCW}
Matthew~E. Peters, Mark Neumann, Mohit Iyyer, Matt Gardner, Christopher Clark,
  Kenton Lee, and Luke~S. Zettlemoyer. 2018.
\newblock Deep contextualized word representations.
\newblock In \emph{NAACL-HLT}.

\bibitem[{Radford et~al.(2018)Radford, Narasimhan, Salimans, and
  Sutskever}]{Radford2018Improving}
Alec Radford, Karthik Narasimhan, Tim Salimans, and Ilya Sutskever. 2018.
\newblock Improving language understanding by generative pre-training.
\newblock \emph{CoRR}.

\bibitem[{Rezende and Mohamed(2015)}]{Rezende2015VariationalIW}
Danilo~Jimenez Rezende and Shakir Mohamed. 2015.
\newblock Variational inference with normalizing flows.
\newblock In \emph{ICML}.

\bibitem[{Socher et~al.(2013)Socher, Perelygin, Wu, Chuang, Manning, Ng, and
  Potts}]{Socher2013RecursiveDM}
Richard Socher, Alex Perelygin, Jean Wu, Jason Chuang, Christopher~D. Manning,
  Andrew Ng, and Christopher Potts. 2013.
\newblock Recursive deep models for semantic compositionality over a sentiment
  treebank.
\newblock In \emph{EMNLP}.

\bibitem[{Tai et~al.(2015)Tai, Socher, and Manning}]{Tai2015ImprovedSR}
Kai~Sheng Tai, Richard Socher, and Christopher~D. Manning. 2015.
\newblock Improved semantic representations from tree-structured long
  short-term memory networks.
\newblock In \emph{ACL}.

\bibitem[{Tang et~al.(2017)Tang, Jin, Fang, Wang, and
  de~Sa}]{Tang2017Rethinking}
Shuai Tang, Hailin Jin, Chen Fang, Zhaowen Wang, and Virginia~R. de~Sa. 2017.
\newblock Rethinking skip-thought: A neighborhood based approach.
\newblock In \emph{RepL4NLP, ACL Workshop}.

\bibitem[{Tang et~al.(2018)Tang, Jin, Fang, Wang, and
  de~Sa}]{Tang2018SpeedingUC}
Shuai Tang, Hailin Jin, Chen Fang, Zhaowen Wang, and Virginia~R. de~Sa. 2018.
\newblock Speeding up context-based sentence representation learning with
  non-autoregressive convolutional decoding.
\newblock In \emph{Rep4NLP@ACL}.

\bibitem[{Wiebe et~al.(2005)Wiebe, Wilson, and Cardie}]{Wiebe2005AnnotatingEO}
Janyce Wiebe, Theresa Wilson, and Claire Cardie. 2005.
\newblock Annotating expressions of opinions and emotions in language.
\newblock \emph{Language Resources and Evaluation}, 39:165--210.

\bibitem[{Wieting et~al.(2015)Wieting, Bansal, Gimpel, and
  Livescu}]{Wieting2015FromPD}
John Wieting, Mohit Bansal, Kevin Gimpel, and Karen Livescu. 2015.
\newblock From paraphrase database to compositional paraphrase model and back.
\newblock \emph{TACL}, 3:345--358.

\bibitem[{Wieting and Gimpel(2017)}]{Wieting2017RevisitingRN}
John Wieting and Kevin Gimpel. 2017.
\newblock Revisiting recurrent networks for paraphrastic sentence embeddings.
\newblock In \emph{ACL}.

\bibitem[{Wieting and Gimpel(2018)}]{Wieting2018Para}
John Wieting and Kevin Gimpel. 2018.
\newblock Paranmt-50m: Pushing the limits of paraphrastic sentence embeddings
  with millions of machine translations.
\newblock In \emph{ACL}.

\bibitem[{Wieting and Kiela(2019)}]{wieting2018no}
John Wieting and Douwe Kiela. 2019.
\newblock \href {https://openreview.net/forum?id=BkgPajAcY7} {No training
  required: Exploring random encoders for sentence classification}.
\newblock In \emph{International Conference on Learning Representations}.

\bibitem[{Zhao et~al.(2015)Zhao, Lu, and Poupart}]{Zhao2015SelfAdaptiveHS}
Han Zhao, Zhengdong Lu, and Pascal Poupart. 2015.
\newblock Self-adaptive hierarchical sentence model.
\newblock In \emph{IJCAI}.

\bibitem[{Zhu et~al.(2015)Zhu, Kiros, Zemel, Salakhutdinov, Urtasun, Torralba,
  and Fidler}]{Zhu2015AligningBA}
Yukun Zhu, Ryan Kiros, Richard~S. Zemel, Ruslan Salakhutdinov, Raquel Urtasun,
  Antonio Torralba, and Sanja Fidler. 2015.
\newblock Aligning books and movies: Towards story-like visual explanations by
  watching movies and reading books.
\newblock \emph{ICCV}, pages 19--27.

\end{thebibliography}
\bibliographystyle{acl_natbib}


\end{document}